\documentclass[]{bytedance_seed}

\usepackage[toc,page,header]{appendix}

\usepackage{minitoc}
\usepackage{makecell}
\usepackage{amssymb}
\usepackage{graphicx}
\usepackage{booktabs}
\usepackage{multirow}
\usepackage{soul}
\usepackage[table,dvipsnames]{xcolor}
\usepackage{wasysym}

\title{Video Generation with Predictive Latents}

\author[1,2]{Yian Zhao}
\author[1,\dagger]{Feng Wang}
\author[1]{Qiushan Guo}
\author[3]{Chang Liu}
\author[3]{Xiangyang Ji}
\author[2]{Jian Zhang}
\author[2]{Jie Chen}

\affiliation[1]{ByteDance Seed}
\affiliation[2]{Peking University}
\affiliation[3]{Tsinghua University}

\contribution[\dagger]{Project lead}

\abstract{
Video Variational Autoencoder (VAE) enables latent video generative modeling by mapping the visual world into compact spatiotemporal latent spaces, improving training efficiency and stability.
While existing video VAEs achieve commendable reconstruction quality, continued optimization of reconstruction does not necessarily translate into improved generative performance. \textit{How to enhance the diffusability of video latents remains a critical and unresolved challenge.}
In this work, inspired by principles of predictive world modeling, we investigate the potential of predictive learning to improve the video generative modeling.
To this end, we introduce a simple and effective predictive reconstruction objective that unifies predictive learning with video reconstruction.
Specifically, we randomly discard future frames and encode only partial past observations, while training the decoder to reconstruct the observed frames and predict future ones simultaneously.
This design encourages the latent space to encode temporally predictive structures and build a more coherent understanding of video dynamics, thereby improving generation quality.
Our model, termed Predictive Video VAE (\textbf{PV-VAE}), achieves superior performance on video generation, with \textbf{52\% faster} convergence and a \textbf{34.42 FVD improvement} over the Wan2.2 VAE on UCF101.
Furthermore, comprehensive analyses demonstrate that PV-VAE not only exhibits favorable scalability, with generative performance improving alongside VAE training, but also yields consistent gains in downstream video understanding, underscoring a latent space that effectively captures temporal coherence and motion priors.
}

\date{\today}
\correspondence{Feng Wang at \url{wangfeng.eve@bytedance.com}}
\checkdata[Project Page]{\url{https://zhao-yian.github.io/PVVAE}}

\begin{document}
\maketitle

%不需要目录就注释掉 注意目录不要和第一页放在一块 要有\newpage
%\newpage
%\tableofcontents
%\newpage

\section{Introduction}

\begin{figure}[!t]
\centering
\includegraphics[width=\linewidth]{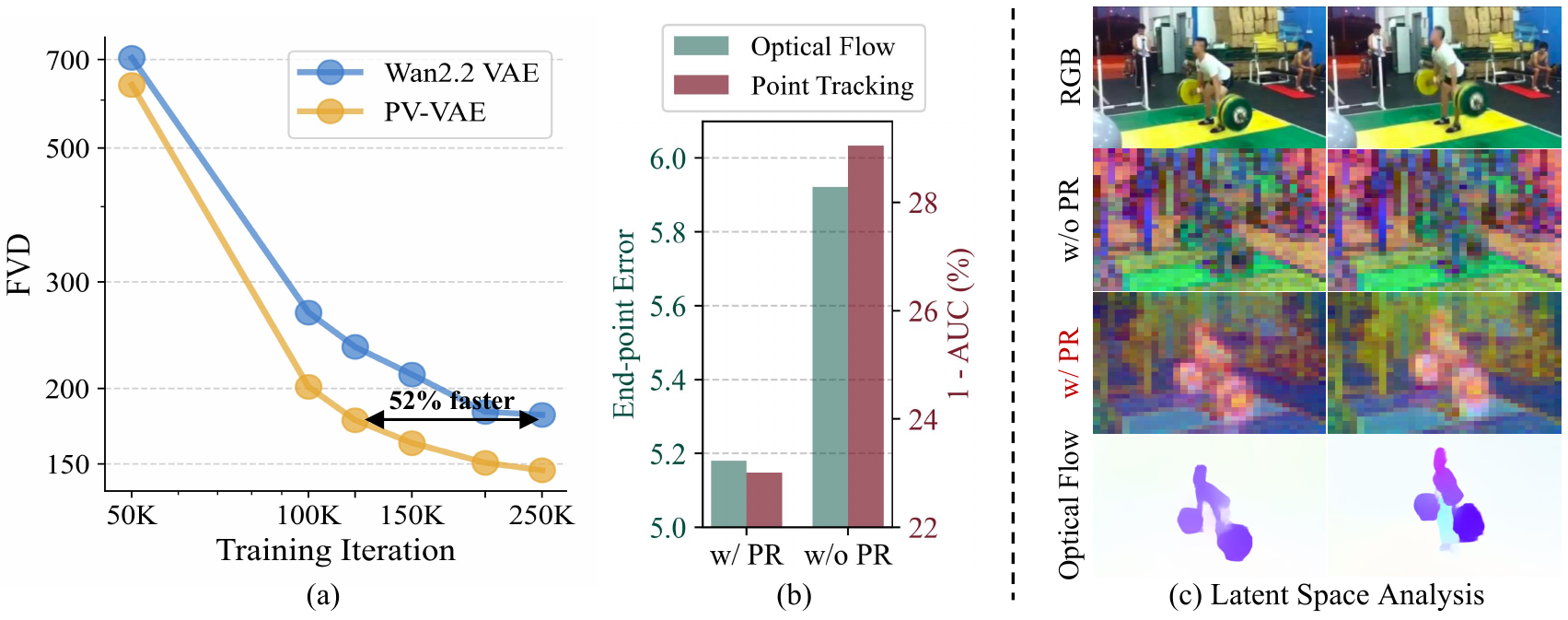} 
\vspace{-0.5em}
\caption{
Our PV-VAE achieves 52\% faster convergence and 34.42 FVD gain over Wan2.2 VAE on UCF101.
Optical flow and point tracking probing tasks show that the Predictive Reconstruction (PR) objective enhances the spatiotemporal understanding of latent space.
Latent visualizations further reveal that PV-VAE captures clear motion-aware structures aligned with video dynamics (visualized via optical flow).
}
\vspace{-0.5em}
\label{fig:first}
\end{figure}

Video generation has achieved extraordinary breakthroughs~\cite{yang2024cogvideox,kong2024hunyuanvideo,wan2025wan,gao2025seedance,seedance2026seedance}, with contemporary models producing content of cinematic brilliance that often surpasses professional-grade cinematography and production standards.
This rapid progress stems from the ability to represent the visual world within compact latent spaces, largely driven by advances in Latent Video Diffusion Models (LVDMs)~\cite{blattmann2023stable} and Video Variational Autoencoders (VAEs)~\cite{pinheiro2021variational}. LVDMs operate not on raw pixels, but on the compact spatiotemporal latent spaces created by video VAEs. These latents not only reduce computational overhead, but more importantly, they provide a structured space for video generative modeling, making video VAEs one of the key components of video generation systems.

The common practice for developing video is to extend well-trained image VAEs and continue training them on video corpora. Modern video VAE~\cite{kong2024hunyuanvideo,wan2025wan} typically adopt CNN-based architectures. They are first trained as 2D image VAEs on large-scale image datasets, after which the 2D convolutions are inflated into 3D causal convolutions to inherit the spatial compression capability~\cite{chen2024od}, followed by video training to achieve joint spatiotemporal compression. 
While existing video VAEs achieve commendable reconstruction quality, continued optimization of reconstruction does not necessarily translate into improved generative performance. 
How to enhance the diffusability~\cite{skorokhodov2025improving} of video latents remains a critical and unresolved challenge.

Different from images, video modeling requires capturing spatiotemporal representations that describe both the visual content and the underlying temporal dynamics from discrete frame sequences. These representations are essential for generating motion-consistent and temporally coherent videos.
Recent studies~\cite{velez2025image,zhu2024exploring} have shown that the representations learned by video generative models yield meaningful results on various video understanding tasks (\textit{e.g.}, depth estimation, tracking, and segmentation), underscoring the crucial role of well-structured video representations in achieving high-quality video generation.
These findings raise a natural question: \textit{what kind of latent spaces enable video generative models to learn temporally structured representations more effectively?}
Inspired by the principle of predictive world modeling~\cite{lecun2022path}, which frames future-state prediction as a powerful means of acquiring temporal and causal structures of videos, we investigate how predictive learning can improve the generative modeling of latent spaces in video VAEs.

Specifically, we introduce a predictive reconstruction objective that unifies video reconstruction with predictive learning.
At each step, we randomly discard future frames, enabling the encoder to observe only partial temporal context, while requiring the decoder to reconstruct the complete video sequence. 
This design forces the model to jointly capture fine-grained visual details and long-term video dynamics, thereby enriching the latent space with robust motion priors that substantially bolster video generation.
Notably, our approach seamlessly integrates into existing video VAE pipelines without altering the original loss composition or introducing additional hyperparameters.
Additionally, to prevent ``copy-shortcut'' from dominating the optimization, a motion-aware objective is incorporated as a targeted constraint, directing the model's attention toward structural motion and fostering more effective predictive learning.

To validate the effectiveness of our approach, we evaluate both class-conditional and unconditional video generation, and show that our model, termed Predictive Video VAE (\textbf{PV-VAE}), consistently achieves notable improvements.
For instance, our PV-VAE achieves 52\% faster convergence and 34.42 FVD improvement over Wan2.2 VAE~\cite{wan2025wan} on UCF101 \cite{soomro2012ucf101} (\textit{cf.} \cref{fig:first}(a)).
To further understand the source of these gains, we examine the learned latent spaces through the lens of diffusion features, which have been shown to serve as reliable intermediate indicators of generative capability~\cite{tang2023emergent,yu2024representation}.
Surprisingly, we find that the diffusion features learned with our PV-VAE exhibit stronger performance across several downstream video understanding tasks, including optical flow estimation~\cite{fleet2006optical}, next-frame prediction~\cite{zhou2020deep}, and point tracking~\cite{doersch2022tap} (\textit{cf.} \cref{fig:first}(b)).
PCA visualizations of the latent space further reveal that PV-VAE captures motion-aware structures that align well with the underlying video dynamics (\textit{cf.} \cref{fig:first}(c)).
These observations indicate that our method strengthens the temporal understanding and motion sensitivity of the learned latent space, leading to improved video generation quality.

In summary, our main contributions are as follows:
\begin{itemize}
\item We investigate the diffusability of video latent spaces and propose a predictive reconstruction objective. By integrating predictive learning into the VAE framework, our method enriches the latent space with robust temporal priors and motion awareness.
\item We develop Predictive Video VAE, which achieves significant improvements across both class-conditional and unconditional video generation, validating the efficacy of our approach.
\item We provide a comprehensive diagnostic of the latent spaces, establishing a clear link between predictive accuracy and generative quality, showing the data scalability of PV-VAE, and demonstrating consistent gains across multiple downstream video understanding tasks.
\end{itemize}
\section{Related Work}

\noindent \textbf{Video VAE.}
Video VAE~\cite{pinheiro2021variational} serves as a fundamental component in modern video generative pipelines.
By employing an encoder–decoder architecture, it maps high-dimensional data into a compact latent space, thereby enhancing the training efficiency and stability of generative models~\cite{rombach2022high}.
Early video generative models~\cite{blattmann2023stable,ma2024latte} directly reused image VAEs to spatially compress individual frames or inserted 1D temporal convolutions into image VAEs to mitigate inter-frame flickering.
Sora~\cite{brooks2024video} first proposed a video compression network for joint spatiotemporal compression to reduce the inference cost.
However, training a video VAE from scratch remains computationally expensive and inefficient. 
To leverage pretrained image VAEs while enabling temporal compression, the community has explored various hybrid designs.
Open-Sora~\cite{zheng2024open} employs a cascade VAE to separately perform spatial and temporal compression.
CV-VAE~\cite{zhao2024cv} introduces latent space alignment between video VAE and image VAE.
OD-VAE~\cite{chen2024od} inflates 2D convolutions of image VAEs into 3D causal convolutions.
% and proposes several variants to balance performance and cost. 
CogVideoX's VAE~\cite{yang2024cogvideox} adopts parallel algorithms for long video processing, while IV-VAE~\cite{wu2025improved} introduces additional channels for temporal compression. 
For improved efficiency, Lite-VAE~\cite{sadat2024litevae} and WF-VAE~\cite{li2025wf} utilize wavelet-based methods, whereas LeanVAE~\cite{cheng2025leanvae} and H3AE~\cite{wu2025h3ae} prioritize structural lightweighting and decoding acceleration.
Additionally, some works~\cite{yu2024efficient,wang2025vidtwin,yin2025deco} decouple motion dynamics from static content to bolster temporal modeling and reduce redundancy.
Recently, many advanced video generative models~\cite{kong2024hunyuanvideo,wan2025wan,gao2025seedance,teng2025magi} have developed unified image-video VAEs.
Despite these advances, little attention has been paid to how the latent spaces can be structured to explicitly benefit video generation.
In this work, we take a step toward addressing this challenge by introducing a predictive reconstruction objective.

\noindent \textbf{Diffusability of latent space.}
Diffusability refers to the suitability of a latent space for the diffusion process. Incorporating structured constraints into the latent space has emerged as a promising approach to improve this.
In the image domain, many frameworks~\cite{yao2025reconstruction,zheng2025diffusion,zhang2025both,leng2025repa} internalize semantic priors from pre-trained encoders (\textit{e.g.}, DINOv2~\cite{oquab2023dinov2}), while VTP~\cite{yao2025towards} advocates for a joint representation-reconstruction learning paradigm. 
Conversely, video-level exploration remains hampered by architectural and computational bottlenecks. 
SSVAE~\cite{liu2025delving} relies on hand-crafted heuristic constraints to shape the latent manifold.
In contrast, our proposed predictive reconstruction encourages the latent space to autonomously capture structured temporal dynamics.

\noindent \textbf{Predictive learning.}
Predictive learning, which aims to predict future states by modeling existing information, has demonstrated powerful representation learning and modeling capabilities across diverse tasks. 
Its applications span from sequence, action, and trajectory prediction~\cite{vu2014predicting,ryoo2011human} to masked language/visual modeling (MLM/MVM)~\cite{devlin2019bert,brown2020language,he2022masked,xie2022simmim}. 
SiameseMAE~\cite{gupta2023siamese} combines predictive learning with masked modeling to learn fine-grained correspondences from randomly sampled video frames. 
JEPA (Joint Embedding Predictive Architecture)~\cite{lecun2022path} further proposes that predictive latent learning serves as a fundamental pathway toward understanding the visual world and constructing world models.
Subsequent works~\cite{assran2023self,bardes2023v,assran2025v,baldassarre2025back} have demonstrated powerful capabilities in visual understanding, prediction, and planning under predictive learning objectives, further validating the effectiveness of this paradigm. 
Most recently, Cambrian-S~\cite{yang2025cambrian} posits predictive sensing as a promising direction for next-generation intelligent agents, offering a proof-of-concept via next-latent-frame prediction.
Building upon these insights, our approach integrates predictive learning with video reconstruction, enabling the model to simultaneously reconstruct visual details and predict future states.
This design enhances the temporal dynamics and motion understanding of latent spaces, thereby facilitating more effective video generative modeling.
\section{Approach}
Our goal is to enhance the diffusability of the latent spaces by jointly learning predictive and reconstruction objectives. 
Let $\mathbf{x} \in \mathbb{R}^{(1+T)\times H\times W\times 3}$ denotes a video clip with $1+T$ frames in pixel space, and $\mathbf{z} \in \mathbb{R}^{(1+t)\times h\times w\times c}$ denotes the sampled video latents.
Here, $p_s = H/h = W/w$ and $p_t = T/t$ are the spatial and temporal compression ratios, and $c$ denotes the latent channel.
The initial extra frame serves to ensure a unified processing pipeline for image ($T{=}0$) and video data, following common practice~\cite{yang2024cogvideox,wan2025wan}.

\begin{figure}[t]
\centering
\includegraphics[width=0.6\linewidth]{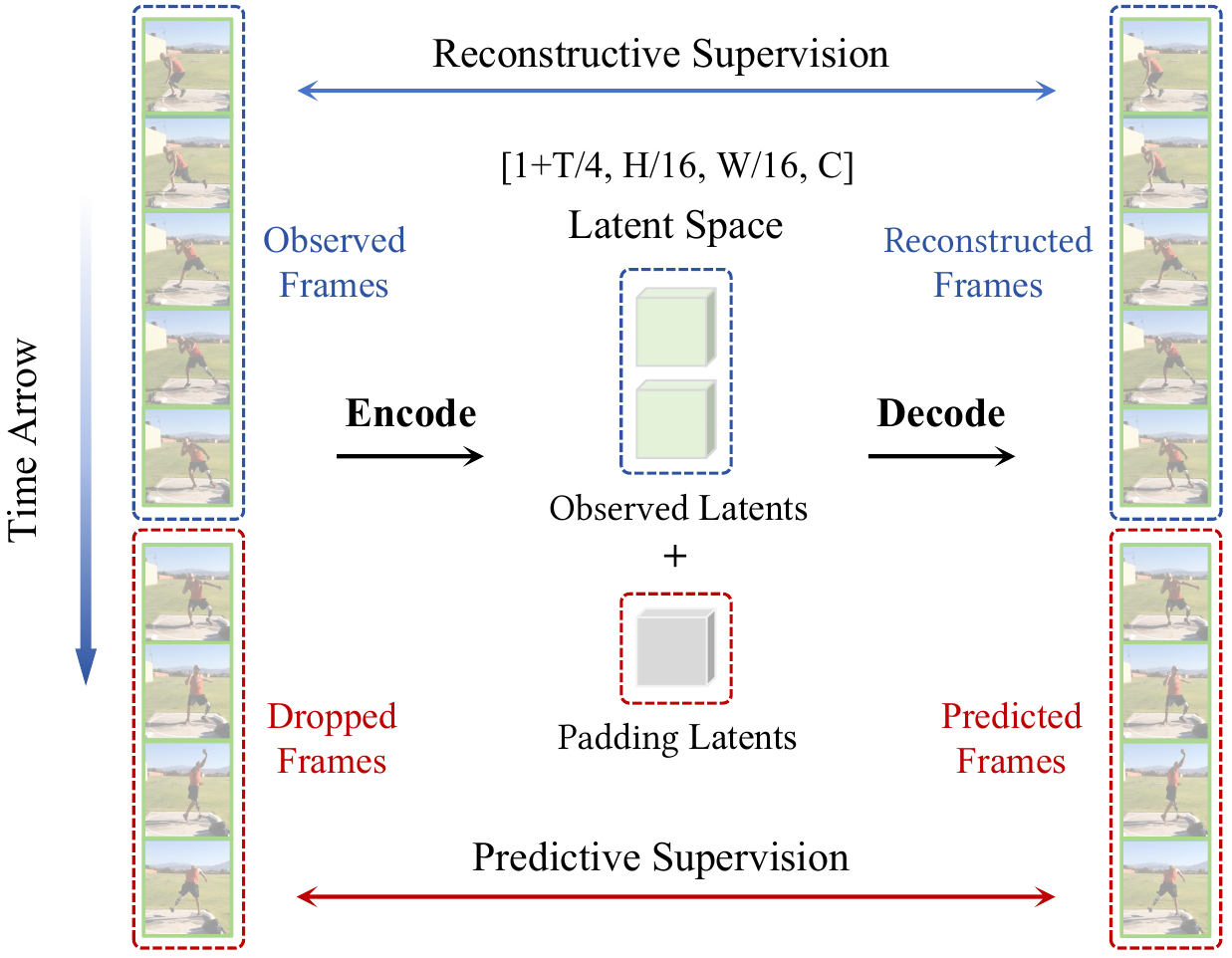}
\caption{
    \textbf{Overall pipeline of the proposed PV-VAE.}
PV-VAE randomly discards future frames and encodes only observed ones. The padded latents are then decoded to reconstruct the full video, enabling the model to learn visual reconstruction and temporal understanding jointly from reconstructive and predictive supervision.
}
\vspace{-0.5em}
\label{fig:overview}
\end{figure}

\subsection{Framework}
\label{subsec:causal}

\noindent \textbf{Integrating predictive learning into reconstruction.}
To incorporate predictive learning, we reformulate the VAE training procedure by introducing a partial-to-complete reconstruction task.
Specifically, we divide the video clip into two parts along the time dimension, denoted as $\mathbf{x} = \langle\mathbf{x}_{obs}, \mathbf{x}_{drop}\rangle$.
The model is trained to reconstruct the entire clip $\mathbf{x}$ conditioned on the observed portion $\mathbf{x}_{obs}$.
At each training step, we first partition the video clip into $G = 1+T/p_t$ groups based on the temporal compression ratio $p_t$, where the first group consists of the first frame, and each subsequent group includes $p_t$ frames.
We then sample the number of dropped groups, $k \sim U\{0, \dots, \lfloor (G-1) \cdot r \rfloor\}$, where $r$ is a predefined maximum dropping ratio.
The retained preceding frames $\mathbf{x}_{obs} \in \mathbb{R}^{(1+T-k\cdot p_t)\times H\times W\times 3}$ are fed into the encoder to obtain the corresponding observed latent $\mathbf{z}_{obs} \in \mathbb{R}^{(G-k)\times h\times w\times c}$.
Given that the decoder shares symmetric spatiotemporal scaling factors with the encoder, it requires a full-length latent sequence to reconstruct the entire video sequence.
As a result, we pad $\mathbf{z}_{obs}$ by temporally concatenating it with padding vectors $\mathbf{z}_{pad} \in \mathbb{R}^{k \times h \times w \times c}$, which are sampled from an uninformative prior (\textit{i.e.}, containing no input information).
This complete latent sequence is passed through the decoder to reconstruct the entire video $\mathbf{x}$.
Since the dropped frames $\mathbf{x}_{drop}$ are entirely withheld from the encoder, the model is compelled to infer the subsequent video evolution from the past observations $\mathbf{x}_{obs}$ and encode this predictive information into its latent spaces.
The overall pipeline of our method is illustrated in \cref{fig:overview}.
Under this learning objective, the model not only learns to reconstruct fine-grained visual details but also develops a deeper understanding of temporal dynamics and motion awareness in videos, thereby improving the latent representations to facilitate better generative modeling.

\noindent \textbf{Model design.}
We implement PV-VAE with 3D causal convolutions, employing $16\times$ spatial and $4\times$ temporal downsampling, with a latent channel dimension of $64$.
For the encoder, we first perform two stages of spatiotemporal downsampling, reducing both the temporal and spatial dimensions by a factor of $4$. Then, while keeping the temporal length fixed, we apply two additional spatial downsampling operations, resulting in an overall $16\times$ spatial reduction.
The decoder is symmetric to the encoder, first conducting two stages of spatial upsampling followed by two stages of spatiotemporal upsampling.

\subsection{Implementation}

\noindent \textbf{Training.}
PV-VAE is first pretrained on multi-resolution image data for 300K steps at resolutions of $256 \times 256$, $384 \times 384$, and $512 \times 512$.
Following this pretraining, it is further trained for 50K steps on video data at $256 \times 256$ and $512 \times 512$ resolutions using the proposed predictive reconstruction objective.
During training, each process randomly samples a varying number of images or videos based on the resolution to maintain a balanced computational load across processes.
Since the decoder requires reconstructing videos from complete video latents during inference, a training–inference gap arises.
To address this issue, we introduce an additional \textit{decoder fine-tuning stage}. Specifically, we freeze the encoder, disable the random frame-dropping operation, and train the decoder for another 50K steps to perform standard video reconstruction.
This stage substantially improves reconstruction quality and provides a stronger foundation for high-fidelity video generation.

\noindent \textbf{Loss functions.}
We adopt a combination of losses commonly used in video VAEs~\cite{yang2024cogvideox,wan2025wan}, including a mean squared error (MSE) loss, a learned perceptual image patch similarity (LPIPS) loss~\cite{zhang2018unreasonable}, an adversarial (GAN) loss~\cite{goodfellow2020generative}, and a KL regularization term.
The GAN loss is activated from step 5,000 during training and remains enabled throughout the entire decoder fine-tuning stage.
To prevent the ``copy-shortcut'' of non-motion regions from dominating the optimization, we incorporate an additional \textit{motion-aware objective}. 
Specifically, the model is required to reconstruct not only the raw pixels but also the temporal differences between adjacent frames. This design effectively filters out static backgrounds and compels the video VAE to prioritize the learning of structural motion and temporal evolution. 
The total loss is formulated as follows:
\begin{equation}
\begin{aligned}
    \mathcal{L}_{total} = & \lambda_{rec}(\mathcal{L}_\text{MSE} + \mathcal{L}_\text{Diff}) + \lambda_{lpips}\mathcal{L}_\text{LPIPS}
    +\lambda_{gan}\mathcal{L}_\text{GAN}+\lambda_{kl}\mathcal{L}_\text{KL},
\end{aligned}
\end{equation}
where each $\lambda$ controls the relative contribution of its corresponding component.
\section{Experiments}
\begin{table*}[!t]
\centering
\setlength{\tabcolsep}{5.5pt}
\renewcommand{\arraystretch}{1}
\caption{
{Comparison of generation performance on the UCF101 and RealEstate10K datasets at 17-frame $256 \times 256$ resolution.}
The best and second-best are indicated in \textbf{bold} and \underline{underlined}.
The notation $tTsScC$ denotes a temporal downsampling factor of $T$, a spatial downsampling factor of $S \times S$, and a latent channel dimension of $C$.
}
\begin{tabular*}{\textwidth}{l c ccc cc ccc}
\toprule
\multirow{2}{*}{\textbf{Method}}  & \multirow{2}{*}{\textbf{\makecell[c]{Latent config}}} & \multicolumn{3}{c}{\textbf{UCF101}}& \multicolumn{2}{c}{\textbf{RealEstate10K}} & \multirow{2}{*}{\textbf{\makecell[c]{TSpeed\\(it/s)}}} & \multirow{2}{*}{\textbf{\makecell[c]{TMem\\(GiB)}}} & \multirow{2}{*}{\textbf{\makecell[c]{Param\\(M)}}} \\
\cmidrule(r){3-5} \cmidrule(r){6-7}
&&\textbf{FVD} $\downarrow$ & \textbf{KVD} $\downarrow$ & \textbf{IS} $\uparrow$ & \textbf{FVD} $\downarrow$ & \textbf{KVD} $\downarrow$ \\
\midrule[1pt]
\midrule
\textbf{CogX-VAE}~\cite{yang2024cogvideox} & t4s8c16 & 176.90 & 16.47 & 64.19 & 94.12 & 10.41 & 0.76 & 85.93 & 216 \\
\textbf{IV-VAE}~\cite{wu2025improved} & t4s8c16 & 175.74 & 22.32 & 64.51 & 92.37 & \underline{8.35} & 1.28 & 88.34 & 242 \\
\textbf{WF-VAE-L}~\cite{li2025wf} & t4s8c16 & 188.19 & 33.01 & \underline{67.49} & 107.26 & 12.56 & 2.52 & 87.36 & 317 \\
\textbf{Hunyuan-VAE}~\cite{kong2024hunyuanvideo} & t4s8c16 & 210.30 & 52.81 & 66.40 & 83.45 & 13.23 & 1.64 & 87.36 & 246 \\
\textbf{Wan2.1 VAE}~\cite{wan2025wan} & t4s8c16 & \underline{167.10} & \textbf{11.54} & 66.04 & {83.84} & 10.64 & 1.88 & 86.44 & 127 \\
\midrule
\textbf{Wan2.2 VAE}~\cite{wan2025wan} & t4s16c48 & 180.79 & 17.80 & {67.32} & 87.15 & 10.11 & {4.96} & {30.90} & 705 \\
\textbf{SSVAE}~\cite{liu2025delving} & t4s16c48 & {168.68} & 19.71 & 66.39 & \underline{79.08} & 8.79 & 3.92 & 34.00 & 315 \\
\rowcolor{blue!8}
\textbf{PV-VAE} & t4s16c64 & \textbf{146.37} & \underline{14.52} & \textbf{69.72} & \textbf{72.50} & \textbf{4.06} & {4.40} & {33.34}  & 661 \\
\bottomrule
\end{tabular*}
\vspace{-0.5em}
\label{tab:generation} 
\end {table*}

\subsection{Experimental setups}
\label{subsec:setup}

\noindent \textbf{Evaluation details.}
We evaluate PV-VAE on three widely used benchmarks: UCF101~\cite{soomro2012ucf101}, RealEstate10K~\cite{zhou2018stereo}, and Kinetics-400~\cite{kay2017kinetics}.
For video generation, we follow prior work~\cite{wu2025improved,chen2024od} and adopt the Latte architecture~\cite{ma2024latte}, a Transformer-based latent diffusion model that supports both unconditional and class-conditional generation.
We use UCF101 for class-conditional generation and RealEstate10K for unconditional generation.
All videos are converted into 17-frame clips at $256 \times 256$ resolution for both training and testing.
For video reconstruction, we randomly sample 2,048 videos from Kinetics-400, which offers better visual quality and higher resolution than UCF-101, making it better suited for assessing reconstruction fidelity. 
We take the first 17 frames of each video and evaluate the model at $256 \times 256$ and $512 \times 512$ resolutions to assess its ability to reconstruct inputs across different spatial scales, which is crucial for video generation.

To assess generation quality, we report Frechet Video Distance (FVD) and Kernel Video Distance (KVD)~\cite{unterthiner2018towards}.
For UCF101, we additionally report the Inception Score (IS)~\cite{saito2020train} computed using the pre-trained C3D model from~\cite{tran2015learning}, following the evaluation protocol of~\cite{chen2024od}.
All metrics are computed over 2048 generated samples.
To assess reconstruction quality, we report reconstruction FVD (rFVD), Peak Signal-to-Noise Ratio (PSNR)~\cite{hore2010image}, Learned Perceptual Image Patch Similarity (LPIPS)~\cite{zhang2018unreasonable}, and Structural Similarity Index Measure (SSIM)~\cite{wang2004image}.
We further measure the training speed (TSpeed) and training memory consumption (TMem) of the generation model along with the inference speed (ISpeed) and inference memory consumption (IMem) of the video VAE.
All speed and memory metrics are measured on 17-frame $256 \times 256$ video clips with a batch size of 4.
To ensure numerical stability, TSpeed and ISpeed are averaged over 100 steps following 50 warm-up steps.

\noindent \textbf{Training details.}
We adopt the AdamW optimizer~\cite{loshchilov2017decoupled} with a base learning rate of 
$5 \times 10^{-5}$.
The learning rate is linearly warmed up and decayed by a factor of $10$ using a cosine schedule.
During random dropping, the first frame is always retained, and the maximum dropping ratio $r$ is set to 1.0.
For generation, we remove the patchify downsampling module of the Latte model~\cite{ma2024latte} to accommodate the higher spatiotemporal compression rate following~\cite{chen2024deep}.
The generation model is trained using rectified flow~\cite{liu2022flow} for 250K steps with a learning rate of $1\times10^{-4}$ and a global batch size of 64, and is evaluated with an Euler sampler using 100 steps.

\subsection{Comparison}
\label{subsec:quantitative}

We compare PV-VAE with several representative video VAEs, including CogVideoX VAE (CogX-VAE)\cite{yang2024cogvideox}, IV-VAE\cite{wu2025improved}, WF-VAE \cite{li2025wf}, HunyuanVideo VAE (Hunyuan-VAE)\cite{kong2024hunyuanvideo}, Wan2.1 VAE, Wan2.2 VAE\cite{wan2025wan}, and SSVAE~\cite{liu2025delving}.

\sethlcolor{blue!8}
\noindent \hl{\textbf{Comparison on generation.}}
\cref{tab:generation} reports the generation performance on UCF101~\cite{soomro2012ucf101} and RealEstate10K~\cite{zhou2018stereo} dataset.
Our PV-VAE achieves the best overall performance among all models.
Notably, compared with video VAEs using a $4 \times 8 \times 8$ downsampling factor, PV-VAE not only attains superior generation quality but also delivers substantial improvements in training speed and memory efficiency.
Taking UCF-101 as an example, PV-VAE outperforms Hunyuan-VAE by 63.93 FVD and achieves a 2.68$\times$ speedup in training while reducing memory consumption by 62\%.
Compared to Wan2.2 VAE / SSVAE, PV-VAE delivers a 34.42 / 22.31 FVD improvement despite using a higher latent-channel dimension.
These results suggest that PV-VAE learns a richer and more structured latent spaces of motion and temporal dynamics, making it highly effective for video generative modeling.

\begin{table*}[!t]
\centering
\setlength{\tabcolsep}{5.2pt}
\renewcommand{\arraystretch}{1}
\caption{
Comparison of reconstruction performance on the Kinetics-400 validation set at different resolutions.
}
\begin{tabular*}{\textwidth}{l cccc cccc cc}
\toprule
\multirow{2}{*}{\textbf{Method}} & \multicolumn{4}{c}{\textbf{$17\times256\times256$}}& \multicolumn{4}{c}{\textbf{$17\times512\times512$}} & \multirow{2}{*}{\textbf{\makecell[c]{ISpeed\\(it/s)}}} & \multirow{2}{*}{\textbf{\makecell[c]{IMem\\(GiB)}}} \\
\cmidrule(r){2-5} \cmidrule(r){6-9}
& \textbf{rFVD}$\downarrow$ & \textbf{PSNR}$\uparrow$ & \textbf{SSIM}$\uparrow$ & \textbf{LPIPS}$\downarrow$ & \textbf{rFVD}$\downarrow$ & \textbf{PSNR}$\uparrow$ & \textbf{SSIM}$\uparrow$ & \textbf{LPIPS}$\downarrow$  \\
\midrule[1pt]
\midrule
\textbf{CogX-VAE}~\cite{yang2024cogvideox} & 4.90 & 33.78 & 0.97 & 0.027 & 1.79 & 36.00 & {0.99} & {0.024} & 0.46 & 13.64 \\
\textbf{IV-VAE}~\cite{wu2025improved} & {2.78} & {34.08} & 0.96 & 0.019 & {0.97} & {37.24} & 0.96 & 0.016 & 0.32 & 5.39 \\
\textbf{WF-VAE-L}~\cite{li2025wf} & 3.06 & 33.48 & 0.96 & 0.023 & 1.08 & 35.93 & 0.96 & {0.023} & 0.87 & 5.00 \\
\textbf{Hunyuan-VAE}~\cite{kong2024hunyuanvideo} & 2.96 & {34.30} & 0.97 & {0.016} & {0.90} & {37.13} & 0.97 & 0.015 & 0.50 & 22.00 \\
\textbf{Wan2.1 VAE}~\cite{wan2025wan} & {2.92} & 33.21 & 0.95 & 0.018 & 1.02 & 36.15 & 0.97 & 0.017 & 0.60 & 6.77 \\
\midrule
\textbf{Wan2.2 VAE}~\cite{wan2025wan} & 3.42 & 33.78 & 0.96 & {0.015} & 1.22 & 36.75 & 0.97 & 0.015 & 0.58 & 9.36 \\
\textbf{SSVAE}~\cite{liu2025delving} & 7.50 & 31.18 & 0.96 & 0.036 & 2.16 & 34.45 & 0.97 & 0.028 & 0.64 &  7.63 \\
\rowcolor{orange!10}
\textbf{PV-VAE} & 3.45 & 32.26 & 0.95 & 0.020 & 1.88 & 35.03 & 0.97 & 0.020 & 0.69 & 7.97 \\
\bottomrule
\end{tabular*}
\label{tab:reconstruction} 
\end {table*}

\begin{figure*}[!t]
\centering
\includegraphics[width=\textwidth]{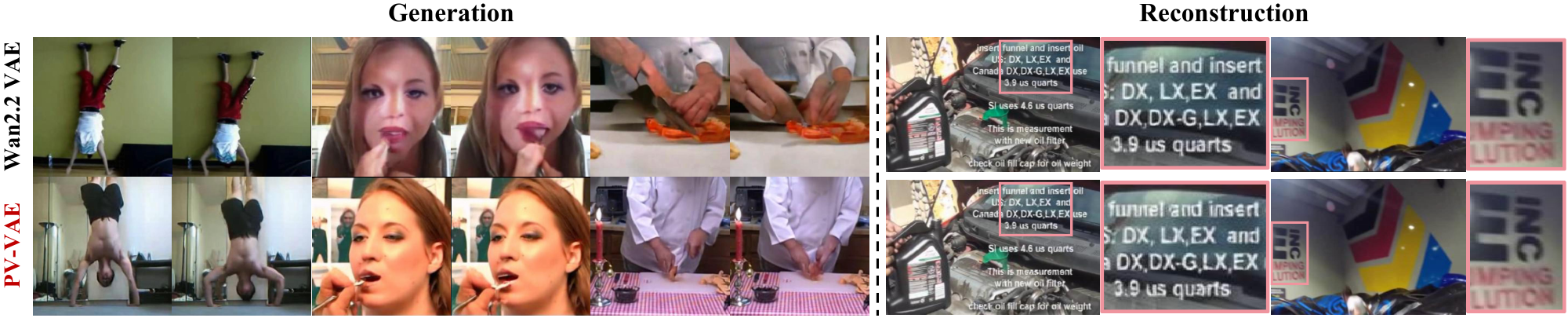}
\vspace{-1em}
\caption{
\sethlcolor{red!8}
\hl{\textbf{Qualitative comparison of generation and reconstruction.}} PV-VAE exhibits enhanced generative quality over Wan2.2 VAE while preserving competitive reconstruction fidelity.
}
\vspace{-1em}
\label{fig:qualitative}
\end{figure*}

\sethlcolor{orange!8}
\noindent \hl{\textbf{Comparison on reconstruction.}}
\cref{tab:reconstruction} presents the reconstruction results on the Kinetics-400~\cite{kay2017kinetics} dataset.
Video VAEs with $4 \times 8 \times 8$ compression typically yield better reconstruction metrics. 
In the context of $4 \times 16 \times 16$ models, PV-VAE delivers reconstruction performance comparable to existing video VAEs. It slightly underperforms relative to Wan2.2 VAE but consistently outperforms SSVAE.
We also test the inference speed and memory consumption of different models at $256 \times 256$ resolution.
Compared to Hunyuan-VAE / Wan2.2 VAE, PV-VAE achieves 38\% / 19\% faster inference while reducing memory consumption by 64\% / 15\%.

\sethlcolor{red!8}
\noindent \hl{\textbf{Qualitative comparison.}}
We further present qualitative comparison results in~\cref{fig:qualitative}. 
Under the same generative training settings, PV-VAE demonstrates superior visual fidelity over the Wan2.2 VAE, while exhibiting fewer motion artifacts and enhanced temporal coherence in video content.
For reconstruction, we select two challenging cases. Notably, PV-VAE exhibits subtle limitations in reconstructing dense text, a performance gap likely stemming from the scarcity of text-heavy samples in our current data distribution~\cite{tong2026scaling}. 
Moving forward, we aim to incorporate more diverse datasets to further elevate the performance upper bound of PV-VAE.

\subsection{Analysis}
\label{subsec:analysis}

To better understand how the proposed predictive reconstruction works, we conduct extensive qualitative and quantitative analyses. Specifically, we dissect the latent space structure via principal component analysis (PCA), demonstrate the correlation between frame prediction accuracy and generation performance, investigate the scaling behaviors, and examine the latent temporal properties. 
Furthermore, we analyze the sources of PV-VAE's performance gains using diffusion features on several downstream video understanding tasks. Finally, we provide visualizations of both reconstruction and future frame prediction to validate the effectiveness of our predictive reconstruction learning.

\begin{figure*}[!t]
\centering
\includegraphics[width=\textwidth]{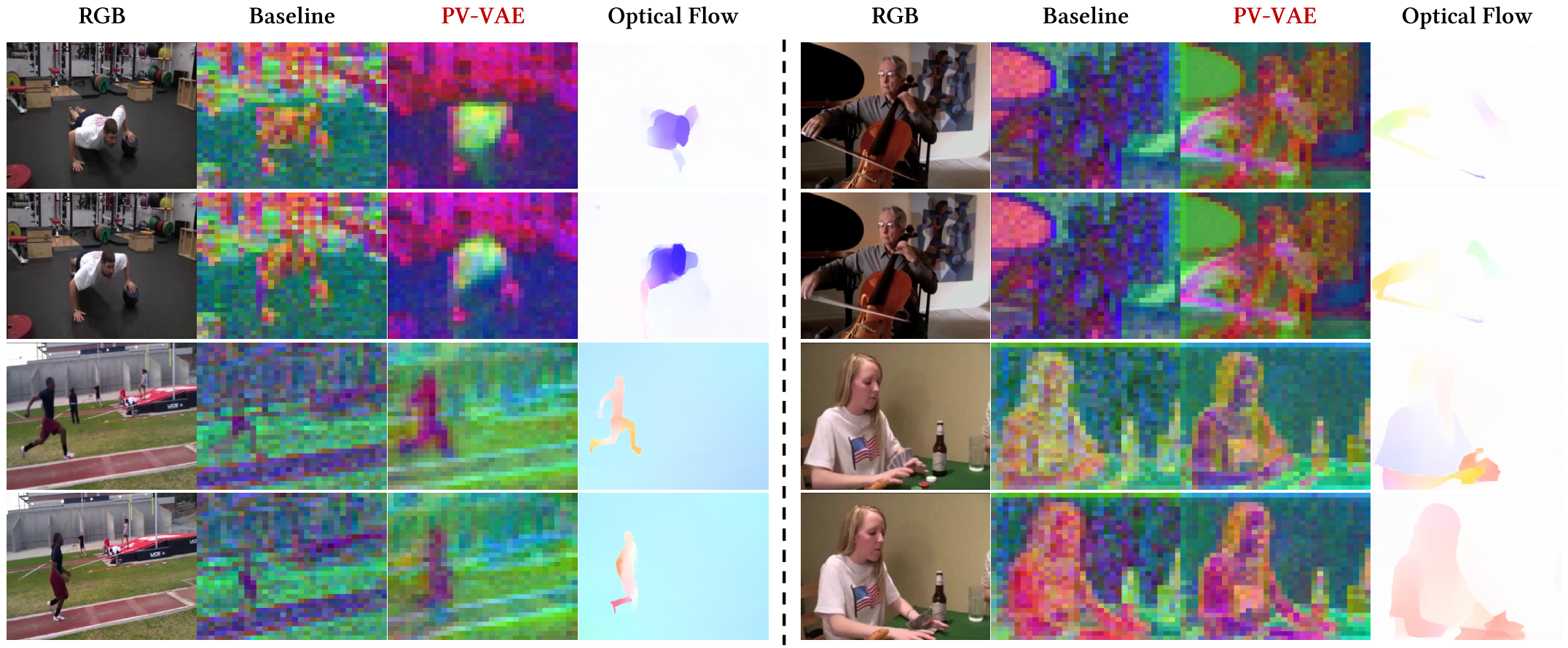}
\vspace{-1.2em}
\caption{
    \sethlcolor{green!10}\textbf{PCA analysis of latent space structure.} PV-VAE exhibits a clear correspondence between latent activations and underlying video motion, with activation patterns strongly aligned with optical flow, indicating that our model effectively concentrates spatiotemporal saliency within its latent representations.
}
\label{fig:pca}
\end{figure*}

\begin{figure}[!t]
\centering
\includegraphics[width=\linewidth]{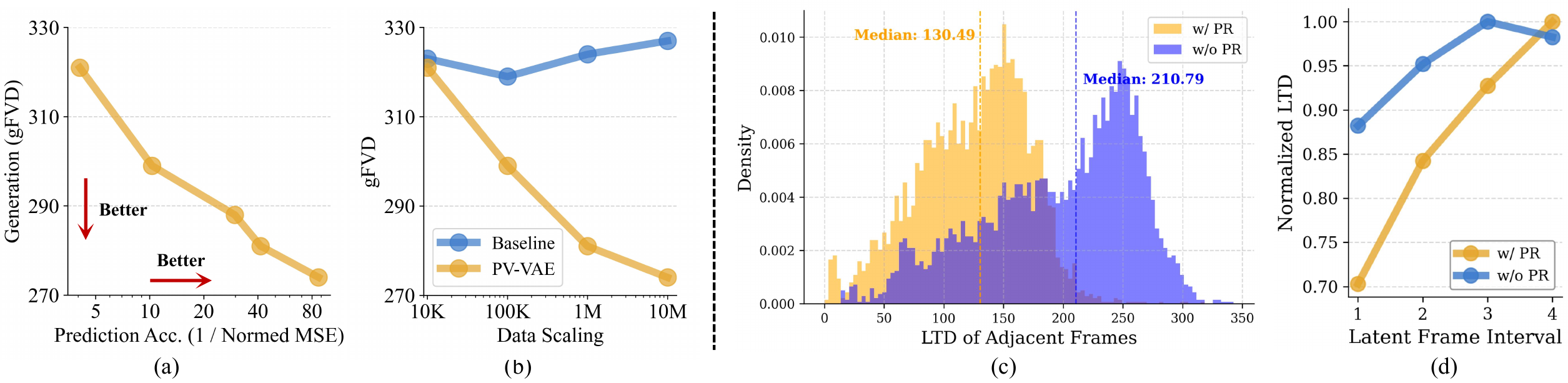}
\vspace{-1.2em}
\caption{
\sethlcolor{yellow!20}
\textbf{(a):} Correlation between generation and prediction accuracy.
\textbf{(b):} Scalability of the predictive reconstruction objective.
\textbf{(c):} Short-term temporal smoothness. PV-VAE achieves higher adjacency coherence than the baseline.
\textbf{(d):} Long-term temporal dynamics. PV-VAE demonstrates a monotonic latent trajectory across expanding frame intervals.
These results collectively validate the effectiveness of predictive reconstruction in imposing structured temporal constraints on video latents.
}
\vspace{-1em}
\label{fig:adj}
\end{figure}

\noindent \textbf{PCA analysis of latent space.}
To investigate the impact of predictive reconstruction on the structure of latent spaces, we perform PCA along the channel dimension of the latents and visualize the top three principal components as RGB images, as shown in \cref{fig:pca}.
We randomly sample several videos from the Kinetics-400~\cite{kay2017kinetics} validation set and compare the PCA visualizations obtained from the baseline model and PV-VAE, alongside the corresponding optical flow computed by RAFT~\cite{teed2020raft}.
For each video, we display two non-adjacent frames to illustrate temporal dynamics.
PV-VAE exhibits a clear correspondence between latents and underlying motion, with activation patterns strongly aligned with optical flow.
Regions with high activation coincide with large motion vectors.
In the left visualizations, the person doing push-ups and the one performing a long jump exhibit noticeably stronger activations than the background.
Similarly, in the right visualizations, the hands of the cello player and the arms and hands of the person playing cards receive higher attention, indicating that the model effectively concentrates spatiotemporal saliency within its latent space.
Moreover, we observe that the background regions with small motion vectors exhibit reduced noise compared to the baseline, suggesting that PV-VAE encourages the latent space to allocate more representational bandwidth to dynamic foregrounds while maintaining smoother, lower-variance representations for static areas.

\noindent \textbf{Correlation study and scaling behaviors.}
To verify the synergy between future prediction and generation, we conduct a correlation study as shown in \cref{fig:adj}(a). 
The results confirm that improved predictive accuracy consistently translates into superior generative performance, justifying our core motivation. On this basis, we further investigate the scaling behavior of PV-VAE in \cref{fig:adj}(b).
We observe consistent performance gains as training data scales, a trend notably absent with the pure reconstruction objective, highlighting the superior scalability of our predictive reconstruction paradigm.

\noindent \textbf{Latent temporal coherence.}
To evaluate temporal coherence, we introduce the Latent Temporal Distance (LTD) metric, computed as the average $L_2$ distance between latents across varying intervals for 1,000 Kinetics-400 validation videos.
As shown in \cref{fig:adj}(c), PV-VAE exhibits a lower median and a sharper histogram peak in adjacent-frame LTD compared to the baseline, suggesting smoother temporal transitions.
Furthermore, as frame intervals grow, PV-VAE demonstrates a consistent monotonic increase in normalized LTD, whereas the baseline lacks this trend, as shown in \cref{fig:adj}(d).
This reveals a smoothly evolving latent trajectory that effectively captures continuous video dynamics, confirming the role of predictive reconstruction in promoting temporal consistency.

\begin{figure*}[!t]
\centering
\includegraphics[width=\textwidth]{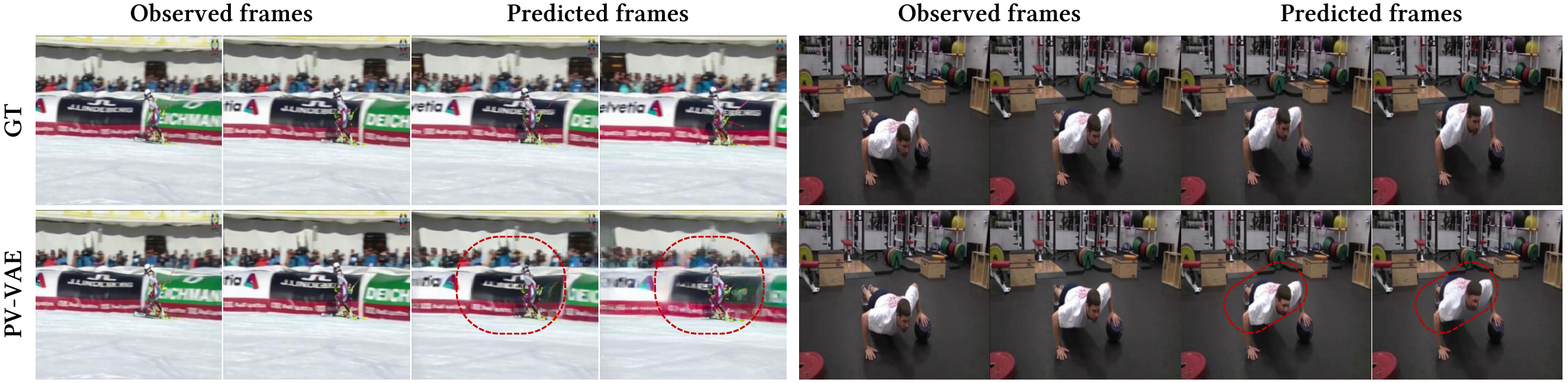}
\caption{
\textbf{Frame Prediction Validation.} PV-VAE generates plausible future frames aligned with underlying temporal evolutions. Red dotted circles highlight shifts in relative object positioning (best viewed under zoom-in).
}
\vspace{-0.5em}
\label{fig:vis}
\end{figure*}

\begin{table}[t]
\centering
\setlength{\tabcolsep}{15pt}
\renewcommand{\arraystretch}{1}
\caption{
    \textbf{Probing results on three video understanding tasks.}
    Compared to the baseline model, PV-VAE achieves consistent gains across all tasks, indicating that our method enhances the learned representations with stronger video understanding.
}
\vspace{-0.5em}
\begin{tabular*}{0.72\linewidth}{l ccc c}
\toprule
\textbf{Method} & \textbf{EPE} $\downarrow$ & \textbf{MSE} $\downarrow$ & \textbf{AUC(\%)} $\uparrow$ \\
\midrule
\textbf{w/o PR} & 5.9223 & 0.0314 & 70.95 \\
\textbf{w/ PR} & \textbf{5.1805} (\textbf{\textcolor[RGB]{65,105,225}{+12.5\%}}) & \textbf{0.0289} (\textbf{\textcolor[RGB]{65,105,225}{+8.0\%}}) & \textbf{76.99} (\textbf{\textcolor[RGB]{65,105,225}{+8.5\%}}) \\
\bottomrule
\end{tabular*}
\vspace{-0.5em}
\label{tab:proxy} 
\end{table}

\noindent \textbf{Probing video understanding in the latent space.}
To dissect the sources of performance gains, we examine the learned latent spaces through the lens of diffusion features across three representative video understanding tasks: optical flow estimation~\cite{fleet2006optical}, next-frame prediction~\cite{zhou2020deep}, and point tracking~\cite{doersch2022tap}, as shown in \cref{tab:proxy}. Features are extracted from the 14th layer (out of 28) of the LVDM for all tasks, with specific configurations detailed below:
\begin{itemize}
\item \textbf{Optical flow estimation:} We utilize the Sintel~\cite{Butler:ECCV:2012} dataset, employing a task-specific decoder with 3D convolutions and pixel-shuffle operations to upsample LVDM features to the original resolution. Performance is quantified by the Average End-Point Error (EPE).
\item \textbf{Next-frame prediction:} Evaluating on Kinetics-400 \cite{kay2017kinetics}, we adapt the flow decoder by adjusting its output channels to three for RGB prediction, reporting the Mean Squared Error (MSE).
\item \textbf{Point tracking:} We evaluate on the TAP-Vid-DAVIS \cite{perazzi2016benchmark} dataset, which contains 30 videos annotated with query points and corresponding ground-truth trajectories. We report the Area Under the Curve (AUC) of tracking accuracy across error thresholds from 0 to 10 pixels.
\end{itemize}
Compared to the baseline model, PV-VAE achieves consistent improvements across all three tasks, demonstrating that its latent space encodes superior video dynamics and motion-aware representations. These findings suggest that the enhanced generative performance stems from a more robust understanding of fundamental video properties, highlighting the potential of predictive reconstruction as a promising direction for video modeling.

\noindent \textbf{Predictive reconstruction visualization.}
Finally, we showcase the prediction capabilities of PV-VAE in \cref{fig:vis}.
For each video, we discard the latter half of the frames and task the model with reconstructing the entire sequence. Two observed frames from the observed half and two predicted frames from the unobserved half are shown.
PV-VAE not only reconstructs the observed frames but also generates plausible future frames that align with the underlying video dynamics. For instance, the model accurately predicts relative spatial shifts between subjects and backgrounds while capturing the temporal progression of actions.
These results provide compelling evidence that PV-VAE effectively captures complex temporal dependencies in video data.

\subsection{Ablation study}
\label{subsec:ablation}

\noindent \textbf{Incremental ablation of PV-VAE.}
We first perform an incremental ablation study to dissect the contribution of each key component in PV-VAE, as summarized in \cref{tab:incremental}. The introduction of predictive reconstruction markedly enhances generation performance, while the motion-aware objective also yields positive gains. Furthermore, decoder fine-tuning significantly improves reconstruction quality, from which the generation metrics also derive a slight benefit.

\noindent \textbf{Maximum dropping ratio.}
We also investigate the impact of the maximum dropping ratio $r$, as shown in \cref{tab:ratio}.
Specifically, we set $r$ to 50\%, 75\%, and 100\%, respectively.
Since reconstruction shows marginal differences following decoder fine-tuning, we focus on comparing the generation performance on the UCF-101 dataset.
The results show that generative performance consistently improves with higher perturbation levels, indicating that stronger predictive regularization encourages the learning of more robust and higher-quality representations.
Therefore, we set the maximum dropping ratio $r$ to 100\% in our training setup.

\begin{table}[t]
\centering
\setlength{\tabcolsep}{16pt}
\renewcommand{\arraystretch}{1}
\caption{
\textbf{Incremental ablation of PV-VAE.} Generation (UCF-101) and reconstruction (Kinetics-400) performance across different configurations. All results are measured at $256 \times 256$ resolution.
}
\vspace{-0.5em}
\begin{tabular*}{\linewidth}{l ccccc}
\toprule
\textbf{Method} & \textbf{gFVD} $\downarrow$ & \textbf{rFVD} $\downarrow$ & \textbf{PSNR} $\uparrow$ & \textbf{SSIM} $\uparrow$ & \textbf{LPIPS} $\downarrow$ \\
\midrule
Baseline & 174.81 & {3.03} & {33.44} & {0.96} & {0.017}  \\
+ Predictive Reconstruction & 156.33 & 5.66 & 31.47 & 0.94 & 0.026 \\
+ Motion-aware Objective & 150.10 & 5.79 & 31.38 & 0.94 & 0.026 \\
+ Decoder Fine-tuning & {146.37} & 3.45 & 32.26 & 0.95 & 0.020 \\
\bottomrule
\end{tabular*}
\label{tab:incremental}
\end{table}

\begin{table}[t]
\centering
\begin{minipage}{0.47\textwidth}
\setlength{\tabcolsep}{15pt}
\renewcommand{\arraystretch}{1}
\caption{
\textbf{Ablation on maximum dropping ratio (MDR).}
}
\vspace{-0.5em}
\begin{tabular}{c ccc}
\toprule
\textbf{MDR} & \textbf{gFVD} $\downarrow$ & \textbf{KVD} $\downarrow$ & \textbf{IS} $\uparrow$ \\
\midrule
% \textbf{0\%} & 174.81 & 19.58 & 68.29 \\
\textbf{50\%} & 159.82 & {14.67} & 69.35 \\
\textbf{75\%} & 154.06 & 16.93 & \textbf{70.27} \\
\textbf{100\%} & \textbf{146.37} & \textbf{14.52} & {69.72} \\
\bottomrule
\end{tabular}
\label{tab:ratio}
\end{minipage}
\hfill %%%%%
\begin{minipage}{0.49\textwidth}
\centering
\setlength{\tabcolsep}{13pt}
\renewcommand{\arraystretch}{1}
\caption{
\textbf{Ablation on padding strategy for latents.}
}
\vspace{-0.5em}
\begin{tabular}{c ccc cccc}
\toprule
\textbf{Padding} & \textbf{gFVD} $\downarrow$ & \textbf{KVD} $\downarrow$ & \textbf{IS} $\uparrow$ \\
\midrule
\textbf{Gaussian} & 150.68 & \textbf{11.87} & 68.01 \\
\textbf{Learnable} & \textbf{146.37} & {14.52} & \textbf{69.72} \\
\bottomrule
\end{tabular}
\label{tab:padding}
\end{minipage}
\vspace{-0.5em}
\end{table}

\begin{table}[t]
\centering
\setlength{\tabcolsep}{10.5pt}
\renewcommand{\arraystretch}{1}
\caption{
\textbf{Performance and efficiency comparison between CNN and Transformer-based video VAEs.}
Results are evaluated at $256\times256$ resolution. $^\clubsuit$ denotes our optimized Transformer-based variant.
}
\vspace{-0.5em}
\begin{tabular*}{\linewidth}{c ccc cccc c}
\toprule
\multirow{2}{*}{\textbf{Padding}} & \multicolumn{3}{c}{\textbf{UCF101}} & \multicolumn{4}{c}{\textbf{Kinetics-400}} & \multirow{2}{*}{\textbf{\makecell[c]{ISpeed\\(it/s)}}} \\
\cmidrule(r){2-4} \cmidrule(r){5-8}
& \textbf{gFVD} $\downarrow$ & \textbf{KVD} $\downarrow$ & \textbf{IS} $\uparrow$ & \textbf{rFVD} $\downarrow$ & \textbf{PSNR} $\uparrow$ & \textbf{SSIM} $\uparrow$ & \textbf{LPIPS} $\downarrow$ & \\
\midrule
\textbf{PV-VAE} & {146.37} & {14.52} & {69.72} & 3.45 & 32.26 & 0.95 & 0.020 & 0.69 \\
\textbf{PV-VAE}$^\clubsuit$ & {178.86} & {20.66} & {69.80} & 4.03 & 33.02 & 0.95 & 0.022 & 1.29 \\
\bottomrule
\end{tabular*}
\vspace{-0.5em}
\label{tab:transformer_cnn}
\end{table}

\noindent \textbf{Padding strategies for latents.}
We further conduct an ablation study on how to pad the video latents. Specifically, we compare two strategies: (i) sampling $\mathbf{z}_{pad}$ from a standard Gaussian distribution, and (ii) using learnable tokens following masked modeling practice~\cite{tong2022videomae}. As shown in \cref{tab:padding}, the learnable tokens yield slightly better generation quality.
\section{Discussion and Conclusion}
\noindent \textbf{Generation \textit{vs.} Reconstruction.}
The trade-off between reconstruction and generation remains central to tokenizer design. 
Rich latent information enhances reconstruction fidelity yet complicates generative modeling, while a highly compressed latents facilitates generation but sacrifices detail.
Previous works typically adopt a low-dimensional latent space to strike a balance.
Recent advancements \cite{yao2025reconstruction,shi2025rectok,yao2025towards} show that high-dimensional latents can actually facilitate generation if they are structured by pre-trained or self-supervised priors.
However, we argue that for video, a structured latent space must encompass both semantics and motion. Our PV-VAE shifts the focus from ``what is in the frame'' to ``what happens next''. By employing a predictive reconstruction objective, we ensure the latent space is motion-aware rather than a mere pixel container.
Notably, this predictive philosophy can be generalized to masked modeling, such as frame infilling or joint spatio-temporal prediction. 
Such self-supervised paradigms could further bolster the robustness and versatility of video latent spaces, a direction we intend to explore in future work.

\noindent \textbf{Advantages of multi-stage training.} 
We introduce an additional decoder fine-tuning stage, a strategic design aimed at further enhancing reconstruction, drawing inspiration from successful approaches in the image domain~\cite{yanglatent,shi2025rectok}.
Our empirical observations reveal that this stage serves as an \textit{effective ``free lunch'' with bounded gains}. 
It consistently refines reconstruction quality while preserving latent diffusability by keeping the encoder frozen. In addition, since the encoder remains unchanged during this phase, the decoder fine-tuning can be conducted in parallel with the diffusion backbone (\textit{e.g.}, DiT) training, thereby substantially accelerating the overall development and iteration efficiency.

\noindent \textbf{Rethinking video VAE Architecture.}
Next, we discuss the architectural design of video VAEs. Despite the dominance of Vision Transformers (ViT)~\cite{dosovitskiy2020image} across most vision tasks, existing video VAEs~\cite{li2025wf,kong2024hunyuanvideo,wan2025wan} still predominantly rely on 3D causal convolutions. 
This reliance prevents video VAEs from leveraging the vast ecosystem of modern techniques optimized for Transformer architectures, while also incurring heavy computational overhead and lacking global modeling capabilities. 
To address these limitations, we explore a minimalist, plain Transformer-based video VAE under the same spatiotemporal compression ratio ($4\times16\times16$, $C{=}64$).
The input is first divided into $4\times16\times16$ spatiotemporal patches, then processed by a stack of Transformer blocks. The decoder directly upsamples the representations back to the original resolution using a pixel-shuffle operation.
Both the encoder and decoder consist of 12 layers each, featuring 16 heads with a 128 $head\_dim$, amounting to a total parameter count of roughly 1.2B.
We compare the reconstruction and generative performance, and inference speed against its CNN-based counterpart, as shown in \cref{tab:transformer_cnn}. Our findings indicate that while the Transformer-based PV-VAE$^\clubsuit$ achieves comparable reconstruction fidelity, its generative capability remains limited.

Despite the current generative gap compared to CNN-based models, we contend that Transformer-based video VAEs hold significant promise for future research, primarily for the following two reasons:
\textit{(i) Computational efficiency}: Despite having a larger number of parameters, the Transformer variant achieves 87\% faster inference speed, effectively mitigating the efficiency bottleneck of current video VAEs, especially when processing long video sequences.
\textit{(ii) Representational flexibility}: 
The Transformer architecture naturally integrates with various video representation learning paradigms, allowing it to flexibly incorporate diverse self-supervised objectives~\cite{tong2022videomae} and further improve latent representations for generative modeling.
In future work, we will further explore optimized architectural configurations and training recipes for Transformer-based video VAEs to fully unlock their latent potential.

\noindent \textbf{Conclusion.}
In this work, we present Predictive Video VAE (PV-VAE), which incorporates a predictive reconstruction objective to jointly optimize visual fidelity and temporal dynamics. This approach yields a more temporally structured and generation-ready video latent space. Extensive downstream evaluations and in-depth analyses demonstrate that PV-VAE effectively captures motion-aware representations, leading to substantial gains in video generation performance. We hope our findings provide meaningful insights for future video VAE research and help push the frontiers of video generative modeling.

\clearpage

\bibliographystyle{plainnat}
\bibliography{main}

@article{wan2025wan,
  title={Wan: Open and advanced large-scale video generative models},
  author={Wan, Team and Wang, Ang and Ai, Baole and Wen, Bin and Mao, Chaojie and Xie, Chen-Wei and Chen, Di and Yu, Feiwu and Zhao, Haiming and Yang, Jianxiao and others},
  journal={arXiv preprint arXiv:2503.20314},
  year={2025}
}

@article{yang2024cogvideox,
  title={Cogvideox: Text-to-video diffusion models with an expert transformer},
  author={Yang, Zhuoyi and Teng, Jiayan and Zheng, Wendi and Ding, Ming and Huang, Shiyu and Xu, Jiazheng and Yang, Yuanming and Hong, Wenyi and Zhang, Xiaohan and Feng, Guanyu and others},
  journal={arXiv preprint arXiv:2408.06072},
  year={2024}
}

@article{kay2017kinetics,
  title={The kinetics human action video dataset},
  author={Kay, Will and Carreira, Joao and Simonyan, Karen and Zhang, Brian and Hillier, Chloe and Vijayanarasimhan, Sudheendra and Viola, Fabio and Green, Tim and Back, Trevor and Natsev, Paul and others},
  journal={arXiv preprint arXiv:1705.06950},
  year={2017}
}

@article{seedance2026seedance,
  title={Seedance 2.0: Advancing Video Generation for World Complexity},
  author={Seedance, Team and Chen, De and Chen, Liyang and Chen, Xin and Chen, Ying and Chen, Zhuo and Chen, Zhuowei and Cheng, Feng and Cheng, Tianheng and Cheng, Yufeng and others},
  journal={arXiv preprint arXiv:2604.14148},
  year={2026}
}

@article{zhou2018stereo,
  title={Stereo magnification: Learning view synthesis using multiplane images},
  author={Zhou, Tinghui and Tucker, Richard and Flynn, John and Fyffe, Graham and Snavely, Noah},
  journal={arXiv preprint arXiv:1805.09817},
  year={2018}
}

@article{soomro2012ucf101,
  title={Ucf101: A dataset of 101 human actions classes from videos in the wild},
  author={Soomro, Khurram and Zamir, Amir Roshan and Shah, Mubarak},
  journal={arXiv preprint arXiv:1212.0402},
  year={2012}
}

@article{ma2024latte,
  title={Latte: Latent diffusion transformer for video generation},
  author={Ma, Xin and Wang, Yaohui and Jia, Gengyun and Chen, Xinyuan and Liu, Ziwei and Li, Yuan-Fang and Chen, Cunjian and Qiao, Yu},
  journal={arXiv preprint arXiv:2401.03048},
  year={2024}
}

@article{liu2022flow,
  title={Flow straight and fast: Learning to generate and transfer data with rectified flow},
  author={Liu, Xingchao and Gong, Chengyue and Liu, Qiang},
  journal={arXiv preprint arXiv:2209.03003},
  year={2022}
}

@inproceedings{tran2015learning,
  title={Learning spatiotemporal features with 3d convolutional networks},
  author={Tran, Du and Bourdev, Lubomir and Fergus, Rob and Torresani, Lorenzo and Paluri, Manohar},
  booktitle={Proceedings of the IEEE international conference on computer vision},
  pages={4489--4497},
  year={2015}
}

@article{unterthiner2018towards,
  title={Towards accurate generative models of video: A new metric \& challenges},
  author={Unterthiner, Thomas and Van Steenkiste, Sjoerd and Kurach, Karol and Marinier, Raphael and Michalski, Marcin and Gelly, Sylvain},
  journal={arXiv preprint arXiv:1812.01717},
  year={2018}
}

@article{saito2020train,
  title={Train sparsely, generate densely: Memory-efficient unsupervised training of high-resolution temporal gan},
  author={Saito, Masaki and Saito, Shunta and Koyama, Masanori and Kobayashi, Sosuke},
  journal={International Journal of Computer Vision},
  volume={128},
  number={10},
  pages={2586--2606},
  year={2020},
  publisher={Springer}
}

@inproceedings{hore2010image,
  title={Image quality metrics: PSNR vs. SSIM},
  author={Hore, Alain and Ziou, Djemel},
  booktitle={2010 20th international conference on pattern recognition},
  pages={2366--2369},
  year={2010},
  organization={IEEE}
}

@article{chen2024deep,
  title={Deep compression autoencoder for efficient high-resolution diffusion models},
  author={Chen, Junyu and Cai, Han and Chen, Junsong and Xie, Enze and Yang, Shang and Tang, Haotian and Li, Muyang and Lu, Yao and Han, Song},
  journal={arXiv preprint arXiv:2410.10733},
  year={2024}
}

@article{wang2004image,
  title={Image quality assessment: from error visibility to structural similarity},
  author={Wang, Zhou and Bovik, Alan C and Sheikh, Hamid R and Simoncelli, Eero P},
  journal={IEEE transactions on image processing},
  volume={13},
  number={4},
  pages={600--612},
  year={2004},
  publisher={IEEE}
}

@inproceedings{zhang2018unreasonable,
  title={The unreasonable effectiveness of deep features as a perceptual metric},
  author={Zhang, Richard and Isola, Phillip and Efros, Alexei A and Shechtman, Eli and Wang, Oliver},
  booktitle={Proceedings of the IEEE conference on computer vision and pattern recognition},
  pages={586--595},
  year={2018}
}

@article{loshchilov2017decoupled,
  title={Decoupled weight decay regularization},
  author={Loshchilov, Ilya and Hutter, Frank},
  journal={arXiv preprint arXiv:1711.05101},
  year={2017}
}

@article{brown2020language,
  title={Language models are few-shot learners},
  author={Brown, Tom and Mann, Benjamin and Ryder, Nick and Subbiah, Melanie and Kaplan, Jared D and Dhariwal, Prafulla and Neelakantan, Arvind and Shyam, Pranav and Sastry, Girish and Askell, Amanda and others},
  journal={Advances in neural information processing systems},
  volume={33},
  pages={1877--1901},
  year={2020}
}

@article{doersch2022tap,
  title={Tap-vid: A benchmark for tracking any point in a video},
  author={Doersch, Carl and Gupta, Ankush and Markeeva, Larisa and Recasens, Adria and Smaira, Lucas and Aytar, Yusuf and Carreira, Joao and Zisserman, Andrew and Yang, Yi},
  journal={Advances in Neural Information Processing Systems},
  volume={35},
  pages={13610--13626},
  year={2022}
}

@article{tong2022videomae,
  title={Videomae: Masked autoencoders are data-efficient learners for self-supervised video pre-training},
  author={Tong, Zhan and Song, Yibing and Wang, Jue and Wang, Limin},
  journal={Advances in neural information processing systems},
  volume={35},
  pages={10078--10093},
  year={2022}
}

@article{velez2025image,
  title={From Image to Video: An Empirical Study of Diffusion Representations},
  author={V{\'e}lez, Pedro and Polan{\'\i}a, Luisa F and Yang, Yi and Zhang, Chuhan and Kabra, Rishabh and Arnab, Anurag and Sajjadi, Mehdi SM},
  journal={arXiv preprint arXiv:2502.07001},
  year={2025}
}

@inproceedings{yao2025reconstruction,
  title={Reconstruction vs. generation: Taming optimization dilemma in latent diffusion models},
  author={Yao, Jingfeng and Yang, Bin and Wang, Xinggang},
  booktitle={Proceedings of the Computer Vision and Pattern Recognition Conference},
  pages={15703--15712},
  year={2025}
}

@article{teng2025magi,
  title={MAGI-1: Autoregressive Video Generation at Scale},
  author={Teng, Hansi and Jia, Hongyu and Sun, Lei and Li, Lingzhi and Li, Maolin and Tang, Mingqiu and Han, Shuai and Zhang, Tianning and Zhang, WQ and Luo, Weifeng and others},
  journal={arXiv preprint arXiv:2505.13211},
  year={2025}
}

@article{yu2024representation,
  title={Representation alignment for generation: Training diffusion transformers is easier than you think},
  author={Yu, Sihyun and Kwak, Sangkyung and Jang, Huiwon and Jeong, Jongheon and Huang, Jonathan and Shin, Jinwoo and Xie, Saining},
  journal={arXiv preprint arXiv:2410.06940},
  year={2024}
}

@article{tang2023emergent,
  title={Emergent correspondence from image diffusion},
  author={Tang, Luming and Jia, Menglin and Wang, Qianqian and Phoo, Cheng Perng and Hariharan, Bharath},
  journal={Advances in Neural Information Processing Systems},
  volume={36},
  pages={1363--1389},
  year={2023}
}

@inproceedings{zhu2024exploring,
  title={Exploring pre-trained text-to-video diffusion models for referring video object segmentation},
  author={Zhu, Zixin and Feng, Xuelu and Chen, Dongdong and Yuan, Junsong and Qiao, Chunming and Hua, Gang},
  booktitle={European Conference on Computer Vision},
  pages={452--469},
  year={2024},
  organization={Springer}
}

@inproceedings{perazzi2016benchmark,
  title={A benchmark dataset and evaluation methodology for video object segmentation},
  author={Perazzi, Federico and Pont-Tuset, Jordi and McWilliams, Brian and Van Gool, Luc and Gross, Markus and Sorkine-Hornung, Alexander},
  booktitle={Proceedings of the IEEE conference on computer vision and pattern recognition},
  pages={724--732},
  year={2016}
}

@inproceedings{he2022masked,
  title={Masked autoencoders are scalable vision learners},
  author={He, Kaiming and Chen, Xinlei and Xie, Saining and Li, Yanghao and Doll{\'a}r, Piotr and Girshick, Ross},
  booktitle={Proceedings of the IEEE/CVF conference on computer vision and pattern recognition},
  pages={16000--16009},
  year={2022}
}

@inproceedings{xie2022simmim,
  title={Simmim: A simple framework for masked image modeling},
  author={Xie, Zhenda and Zhang, Zheng and Cao, Yue and Lin, Yutong and Bao, Jianmin and Yao, Zhuliang and Dai, Qi and Hu, Han},
  booktitle={Proceedings of the IEEE/CVF conference on computer vision and pattern recognition},
  pages={9653--9663},
  year={2022}
}

@inproceedings{vu2014predicting,
  title={Predicting actions from static scenes},
  author={Vu, Tuan-Hung and Olsson, Catherine and Laptev, Ivan and Oliva, Aude and Sivic, Josef},
  booktitle={European Conference on Computer Vision},
  pages={421--436},
  year={2014},
  organization={Springer}
}

@inproceedings{ryoo2011human,
  title={Human activity prediction: Early recognition of ongoing activities from streaming videos},
  author={Ryoo, Michael S},
  booktitle={2011 international conference on computer vision},
  pages={1036--1043},
  year={2011},
  organization={IEEE}
}

@inproceedings{teed2020raft,
  title={Raft: Recurrent all-pairs field transforms for optical flow},
  author={Teed, Zachary and Deng, Jia},
  booktitle={European conference on computer vision},
  pages={402--419},
  year={2020},
  organization={Springer}
}

@article{skorokhodov2025improving,
  title={Improving the diffusability of autoencoders},
  author={Skorokhodov, Ivan and Girish, Sharath and Hu, Benran and Menapace, Willi and Li, Yanyu and Abdal, Rameen and Tulyakov, Sergey and Siarohin, Aliaksandr},
  journal={arXiv preprint arXiv:2502.14831},
  year={2025}
}

@article{gupta2023siamese,
  title={Siamese masked autoencoders},
  author={Gupta, Agrim and Wu, Jiajun and Deng, Jia and Li, Fei-Fei},
  journal={Advances in Neural Information Processing Systems},
  volume={36},
  pages={40676--40693},
  year={2023}
}

@article{lecun2022path,
  title={A path towards autonomous machine intelligence version 0.9. 2, 2022-06-27},
  author={LeCun, Yann},
  journal={Open Review},
  volume={62},
  number={1},
  pages={1--62},
  year={2022}
}

@inproceedings{assran2023self,
  title={Self-supervised learning from images with a joint-embedding predictive architecture},
  author={Assran, Mahmoud and Duval, Quentin and Misra, Ishan and Bojanowski, Piotr and Vincent, Pascal and Rabbat, Michael and LeCun, Yann and Ballas, Nicolas},
  booktitle={Proceedings of the IEEE/CVF Conference on Computer Vision and Pattern Recognition},
  pages={15619--15629},
  year={2023}
}

@article{bardes2023v,
  title={V-jepa: Latent video prediction for visual representation learning},
  author={Bardes, Adrien and Garrido, Quentin and Ponce, Jean and Chen, Xinlei and Rabbat, Michael and LeCun, Yann and Assran, Mido and Ballas, Nicolas},
  year={2023}
}

@article{assran2025v,
  title={V-jepa 2: Self-supervised video models enable understanding, prediction and planning},
  author={Assran, Mido and Bardes, Adrien and Fan, David and Garrido, Quentin and Howes, Russell and Muckley, Matthew and Rizvi, Ammar and Roberts, Claire and Sinha, Koustuv and Zholus, Artem and others},
  journal={arXiv preprint arXiv:2506.09985},
  year={2025}
}

@article{baldassarre2025back,
  title={Back to the features: Dino as a foundation for video world models},
  author={Baldassarre, Federico and Szafraniec, Marc and Terver, Basile and Khalidov, Vasil and Massa, Francisco and LeCun, Yann and Labatut, Patrick and Seitzer, Maximilian and Bojanowski, Piotr},
  journal={arXiv preprint arXiv:2507.19468},
  year={2025}
}

@article{goodfellow2020generative,
  title={Generative adversarial networks},
  author={Goodfellow, Ian and Pouget-Abadie, Jean and Mirza, Mehdi and Xu, Bing and Warde-Farley, David and Ozair, Sherjil and Courville, Aaron and Bengio, Yoshua},
  journal={Communications of the ACM},
  volume={63},
  number={11},
  pages={139--144},
  year={2020},
  publisher={ACM New York, NY, USA}
}

@article{yang2025cambrian,
  title={Cambrian-S: Towards Spatial Supersensing in Video},
  author={Yang, Shusheng and Yang, Jihan and Huang, Pinzhi and Brown, Ellis and Yang, Zihao and Yu, Yue and Tong, Shengbang and Zheng, Zihan and Xu, Yifan and Wang, Muhan and others},
  journal={arXiv preprint arXiv:2511.04670},
  year={2025}
}

@inproceedings{devlin2019bert,
  title={Bert: Pre-training of deep bidirectional transformers for language understanding},
  author={Devlin, Jacob and Chang, Ming-Wei and Lee, Kenton and Toutanova, Kristina},
  booktitle={Proceedings of the 2019 conference of the North American chapter of the association for computational linguistics: human language technologies, volume 1 (long and short papers)},
  pages={4171--4186},
  year={2019}
}

@article{zhou2020deep,
  title={Deep learning in next-frame prediction: A benchmark review},
  author={Zhou, Yufan and Dong, Haiwei and El Saddik, Abdulmotaleb},
  journal={IEEE Access},
  volume={8},
  pages={69273--69283},
  year={2020},
  publisher={IEEE}
}

@inproceedings{Butler:ECCV:2012,
title = {A naturalistic open source movie for optical flow evaluation},
author = {Butler, D. J. and Wulff, J. and Stanley, G. B. and Black, M. J.},
booktitle = {European Conf. on Computer Vision (ECCV)},
editor = {{A. Fitzgibbon et al. (Eds.)}},
publisher = {Springer-Verlag},
series = {Part IV, LNCS 7577},
month = oct,
pages = {611--625},
year = {2012}
}

@incollection{fleet2006optical,
  title={Optical flow estimation},
  author={Fleet, David and Weiss, Yair},
  booktitle={Handbook of mathematical models in computer vision},
  pages={237--257},
  year={2006},
  publisher={Springer}
}

@article{blattmann2023stable,
  title={Stable video diffusion: Scaling latent video diffusion models to large datasets},
  author={Blattmann, Andreas and Dockhorn, Tim and Kulal, Sumith and Mendelevitch, Daniel and Kilian, Maciej and Lorenz, Dominik and Levi, Yam and English, Zion and Voleti, Vikram and Letts, Adam and others},
  journal={arXiv preprint arXiv:2311.15127},
  year={2023}
}

@article{brooks2024video,
  title={Video generation models as world simulators},
  author={Brooks, Tim and Peebles, Bill and Holmes, Connor and DePue, Will and Guo, Yufei and Jing, Li and Schnurr, David and Taylor, Joe and Luhman, Troy and Luhman, Eric and others},
  journal={OpenAI Blog},
  volume={1},
  number={8},
  pages={1},
  year={2024}
}

@article{zheng2024open,
  title={Open-sora: Democratizing efficient video production for all},
  author={Zheng, Zangwei and Peng, Xiangyu and Yang, Tianji and Shen, Chenhui and Li, Shenggui and Liu, Hongxin and Zhou, Yukun and Li, Tianyi and You, Yang},
  journal={arXiv preprint arXiv:2412.20404},
  year={2024}
}

@inproceedings{rombach2022high,
  title={High-resolution image synthesis with latent diffusion models},
  author={Rombach, Robin and Blattmann, Andreas and Lorenz, Dominik and Esser, Patrick and Ommer, Bj{\"o}rn},
  booktitle={Proceedings of the IEEE/CVF conference on computer vision and pattern recognition},
  pages={10684--10695},
  year={2022}
}

@article{zhao2024cv,
  title={Cv-vae: A compatible video vae for latent generative video models},
  author={Zhao, Sijie and Zhang, Yong and Cun, Xiaodong and Yang, Shaoshu and Niu, Muyao and Li, Xiaoyu and Hu, Wenbo and Shan, Ying},
  journal={Advances in Neural Information Processing Systems},
  volume={37},
  pages={12847--12871},
  year={2024}
}

@article{chen2024od,
  title={Od-vae: An omni-dimensional video compressor for improving latent video diffusion model},
  author={Chen, Liuhan and Li, Zongjian and Lin, Bin and Zhu, Bin and Wang, Qian and Yuan, Shenghai and Zhou, Xing and Cheng, Xinhua and Yuan, Li},
  journal={arXiv preprint arXiv:2409.01199},
  year={2024}
}

@inproceedings{wu2025improved,
  title={Improved video vae for latent video diffusion model},
  author={Wu, Pingyu and Zhu, Kai and Liu, Yu and Zhao, Liming and Zhai, Wei and Cao, Yang and Zha, Zheng-Jun},
  booktitle={Proceedings of the Computer Vision and Pattern Recognition Conference},
  pages={18124--18133},
  year={2025}
}

@article{sadat2024litevae,
  title={Litevae: Lightweight and efficient variational autoencoders for latent diffusion models},
  author={Sadat, Seyedmorteza and Buhmann, Jakob and Bradley, Derek and Hilliges, Otmar and Weber, Romann M},
  journal={Advances in Neural Information Processing Systems},
  volume={37},
  pages={3907--3936},
  year={2024}
}

@inproceedings{li2025wf,
  title={Wf-vae: Enhancing video vae by wavelet-driven energy flow for latent video diffusion model},
  author={Li, Zongjian and Lin, Bin and Ye, Yang and Chen, Liuhan and Cheng, Xinhua and Yuan, Shenghai and Yuan, Li},
  booktitle={Proceedings of the Computer Vision and Pattern Recognition Conference},
  pages={17778--17788},
  year={2025}
}

@article{kong2024hunyuanvideo,
  title={Hunyuanvideo: A systematic framework for large video generative models},
  author={Kong, Weijie and Tian, Qi and Zhang, Zijian and Min, Rox and Dai, Zuozhuo and Zhou, Jin and Xiong, Jiangfeng and Li, Xin and Wu, Bo and Zhang, Jianwei and others},
  journal={arXiv preprint arXiv:2412.03603},
  year={2024}
}

@article{gao2025seedance,
  title={Seedance 1.0: Exploring the Boundaries of Video Generation Models},
  author={Gao, Yu and Guo, Haoyuan and Hoang, Tuyen and Huang, Weilin and Jiang, Lu and Kong, Fangyuan and Li, Huixia and Li, Jiashi and Li, Liang and Li, Xiaojie and others},
  journal={arXiv preprint arXiv:2506.09113},
  year={2025}
}

@incollection{pinheiro2021variational,
  title={Variational autoencoder},
  author={Pinheiro Cinelli, Lucas and Ara{\'u}jo Marins, Matheus and Barros da Silva, Eduardo Ant{\'u}nio and Lima Netto, S{\'e}rgio},
  booktitle={Variational methods for machine learning with applications to deep networks},
  pages={111--149},
  year={2021},
  publisher={Springer}
}

@article{shi2025rectok,
  title={RecTok: Reconstruction Distillation along Rectified Flow},
  author={Shi, Qingyu and Wu, Size and Bai, Jinbin and Yu, Kaidong and Wang, Yujing and Tong, Yunhai and Li, Xiangtai and Li, Xuelong},
  journal={arXiv preprint arXiv:2512.13421},
  year={2025}
}

@article{yao2025towards,
  title={Towards Scalable Pre-training of Visual Tokenizers for Generation},
  author={Yao, Jingfeng and Song, Yuda and Zhou, Yucong and Wang, Xinggang},
  journal={arXiv preprint arXiv:2512.13687},
  year={2025}
}

@inproceedings{yanglatent,
  title={Latent Denoising Makes Good Tokenizers},
  author={Yang, Jiawei and Li, Tianhong and Fan, Lijie and Tian, Yonglong and Wang, Yue},
  booktitle={The Fourteenth International Conference on Learning Representations},
  year={2026}
}

@article{dosovitskiy2020image,
  title={An image is worth 16x16 words: Transformers for image recognition at scale},
  author={Dosovitskiy, Alexey and Beyer, Lucas and Kolesnikov, Alexander and Weissenborn, Dirk and Zhai, Xiaohua and Unterthiner, Thomas and Dehghani, Mostafa and Minderer, Matthias and Heigold, Georg and Gelly, Sylvain and others},
  journal={arXiv preprint arXiv:2010.11929},
  year={2020}
}

@article{tong2026scaling,
  title={Scaling Text-to-Image Diffusion Transformers with Representation Autoencoders},
  author={Tong, Shengbang and Zheng, Boyang and Wang, Ziteng and Tang, Bingda and Ma, Nanye and Brown, Ellis and Yang, Jihan and Fergus, Rob and LeCun, Yann and Xie, Saining},
  journal={arXiv preprint arXiv:2601.16208},
  year={2026}
}

@article{yin2025deco,
  title={DeCo-VAE: Learning Compact Latents for Video Reconstruction via Decoupled Representation},
  author={Yin, Xiangchen and Yuan, Jiahui and Hu, Zhangchi and Sun, Wenzhang and Chen, Jie and Qiao, Xiaozhen and Li, Hao and Sun, Xiaoyan},
  journal={arXiv preprint arXiv:2511.14530},
  year={2025}
}

@inproceedings{wang2025vidtwin,
  title={Vidtwin: Video vae with decoupled structure and dynamics},
  author={Wang, Yuchi and Guo, Junliang and Xie, Xinyi and He, Tianyu and Sun, Xu and Bian, Jiang},
  booktitle={Proceedings of the Computer Vision and Pattern Recognition Conference},
  pages={22922--22932},
  year={2025}
}

@article{yu2024efficient,
  title={Efficient video diffusion models via content-frame motion-latent decomposition},
  author={Yu, Sihyun and Nie, Weili and Huang, De-An and Li, Boyi and Shin, Jinwoo and Anandkumar, Anima},
  journal={arXiv preprint arXiv:2403.14148},
  year={2024}
}

@article{zheng2025diffusion,
  title={Diffusion transformers with representation autoencoders},
  author={Zheng, Boyang and Ma, Nanye and Tong, Shengbang and Xie, Saining},
  journal={arXiv preprint arXiv:2510.11690},
  year={2025}
}

@article{oquab2023dinov2,
  title={Dinov2: Learning robust visual features without supervision},
  author={Oquab, Maxime and Darcet, Timoth{\'e}e and Moutakanni, Th{\'e}o and Vo, Huy and Szafraniec, Marc and Khalidov, Vasil and Fernandez, Pierre and Haziza, Daniel and Massa, Francisco and El-Nouby, Alaaeldin and others},
  journal={arXiv preprint arXiv:2304.07193},
  year={2023}
}

@article{zhang2025both,
  title={Both Semantics and Reconstruction Matter: Making Representation Encoders Ready for Text-to-Image Generation and Editing},
  author={Zhang, Shilong and Zhang, He and Zhang, Zhifei and Ge, Chongjian and Xue, Shuchen and Liu, Shaoteng and Ren, Mengwei and Kim, Soo Ye and Zhou, Yuqian and Liu, Qing and others},
  journal={arXiv preprint arXiv:2512.17909},
  year={2025}
}

@article{liu2025delving,
  title={Delving into latent spectral biasing of video vaes for superior diffusability},
  author={Liu, Shizhan and Deng, Xinran and Yang, Zhuoyi and Teng, Jiayan and Gu, Xiaotao and Tang, Jie},
  journal={arXiv preprint arXiv:2512.05394},
  year={2025}
}

@inproceedings{leng2025repa,
  title={Repa-e: Unlocking vae for end-to-end tuning of latent diffusion transformers},
  author={Leng, Xingjian and Singh, Jaskirat and Hou, Yunzhong and Xing, Zhenchang and Xie, Saining and Zheng, Liang},
  booktitle={Proceedings of the IEEE/CVF International Conference on Computer Vision},
  pages={18262--18272},
  year={2025}
}

@article{wu2025h3ae,
  title={H3ae: High compression, high speed, and high quality autoencoder for video diffusion models},
  author={Wu, Yushu and Li, Yanyu and Skorokhodov, Ivan and Kag, Anil and Menapace, Willi and Girish, Sharath and Siarohin, Aliaksandr and Wang, Yanzhi and Tulyakov, Sergey},
  journal={arXiv preprint arXiv:2504.10567},
  year={2025}
}

@inproceedings{cheng2025leanvae,
  title={Leanvae: An ultra-efficient reconstruction vae for video diffusion models},
  author={Cheng, Yu and Yuan, Fajie},
  booktitle={Proceedings of the IEEE/CVF International Conference on Computer Vision},
  pages={15692--15702},
  year={2025}
}

% \clearpage
% \beginappendix
% \input{sections/appendix}

\end{document}